%% file: eccv2020submission.tex
\documentclass[runningheads]{llncs}
\usepackage{graphicx}
\usepackage{bmpsize}
\usepackage{comment}
\usepackage{amsmath,amssymb} 
\usepackage[dvipsnames,table]{xcolor}
\usepackage{dsfont}
\usepackage{pifont}
\usepackage{subcaption}
\usepackage{authblk}
\usepackage{epsfig}
\usepackage{booktabs}
\usepackage{xspace}
\usepackage{multirow}
\usepackage{wrapfig}
\usepackage{tabularx,booktabs}
\usepackage{caption}
\usepackage{adjustbox}
\usepackage{framed}
\usepackage{algorithm,algpseudocode}
\usepackage{bibunits}

\defaultbibliographystyle{splncs04}
\defaultbibliography{5-egbib.bib}

\newcommand{\etal}{\textit{et al.}}
\newcommand{\ie}{\textit{i}.\textit{e}.\xspace}
\newcommand{\eg}{\textit{e}.\textit{g}.\xspace}

\DeclareMathOperator*{\argmin}{arg\,min}

\newtheorem{property_}{Property}

\newenvironment{proof-sketch}{\noindent{\bf Sketch of Proof}\hspace*{1em}}{\qed\bigskip}
\newenvironment{proof1}{\noindent{\bf Proof for the shortest path property}\hspace*{1em}}{\qed\bigskip}
\newenvironment{proof2}{\noindent{\bf Proof for the assignment invariance property}\hspace*{1em}}{\qed\bigskip}
\newenvironment{proof3}{\noindent{\bf Proof for linearity}\hspace*{1em}}{\qed\bigskip}
\definecolor{mygray}{gray}{0.95}

\begin{document}
\pagestyle{headings}
\mainmatter
\def\ECCVSubNumber{378}  

\title{PointMixup: Augmentation for Point Clouds} 

\titlerunning{PointMixup: Augmentation for Point Clouds.}
%
\author{Yunlu Chen$^\star$\inst{1}\and
Vincent Tao Hu\thanks{Equal contribution. \email{\{y.chen3, t.hu\}@uva.nl}}\inst{1} \and Efstratios Gavves\inst{1} \and Thomas Mensink\inst{2,1} \and
Pascal Mettes\inst{1}  \and Pengwan Yang\inst{1,3} \and  Cees G. M. Snoek\inst{1}}
\authorrunning{Y. Chen et al.}
%
\institute{University of Amsterdam \and
Google Research, Amsterdam \and Peking University}

\maketitle

\begin{abstract}
This paper introduces data augmentation for point clouds by interpolation between examples. Data augmentation by interpolation has shown to be a simple and effective approach in the image domain. Such a mixup is however not directly transferable to point clouds, as we do not have a one-to-one correspondence between the points of two different objects. In this paper, we define data augmentation between point clouds as a shortest path linear interpolation. To that end, we introduce PointMixup, an interpolation method that generates new  examples through an optimal assignment of the path function between two point clouds. We prove that our PointMixup finds the shortest path between two point clouds and that the interpolation is assignment invariant and linear. With the definition of interpolation, PointMixup allows to introduce strong interpolation-based regularizers such as mixup and manifold mixup to the point cloud domain. Experimentally, we show the potential of PointMixup for point cloud classification, especially when examples are scarce, as well as increased robustness to noise and geometric transformations to points. The code for PointMixup and the experimental details are publicly available\footnote[1]{Code is available at:  \url{https://github.com/yunlu-chen/PointMixup/}}.
\keywords{interpolation, point cloud classification, data augmentation}
\end{abstract}

\begin{bibunit}
\input{0-intro.tex}
\input{1-relate.tex}

\input{2-method-v3}
\input{3-exp-v3.tex}
\input{4-conclude.tex}

\input{eccv2020submission.bbl}
\end{bibunit}

%
%

\newpage

\begin{bibunit}

\appendix
\addcontentsline{toc}{section}{Appendices}
\section*{APPENDICES}
\input{6-supp}
\end{bibunit}

\end{document}

%% file: 0-intro.tex
\section{Introduction}

The goal of this paper is to classify a cloud of points into their semantic category, be it an airplane, a bathtub or a chair. Point cloud classification is challenging, as they are sets and hence invariant to point permutations. Building on the pioneering PointNet by Qi \etal~\cite{qi2017pointnet}, multiple works have proposed deep learning solutions to point cloud classification~\cite{qi2017pointnet++,li2018pointcnn,dgcnn,wu2019pointconv,zhang2019shellnet,thomas2019kpconv}. Given the progress in point cloud network architectures, as well as the importance of data augmentation in improving classification accuracy and robustness, we study how could data augmentation be naturally extended to support also point cloud data, especially considering the often smaller size of point clouds datasets (\eg ModelNet40~\cite{modelnet40}).
In this work, we propose point cloud data augmentation by interpolation of existing training point clouds.

To perform data augmentation by interpolation, we take inspiration from augmentation in the image domain. Several works have shown that generating new training examples, by interpolating images and their corresponding labels, leads to improved network regularization and generalization, \eg,~\cite{adamixup,tokozume2018between,mixup,manifoldmixup}. Such a mixup is feasible in the image domain, due to the regular structure of images and one-to-one correspondences between pixels. However, this setup does not generalize to the point cloud domain, since there is no one-to-one correspondence and ordering between points. To that end, we seek to find a method to enable interpolation between permutation invariant point sets.

\begin{figure}[t]
\centering
\begin{minipage}[]{1.0\textwidth}
\centering
\includegraphics[width=\textwidth]{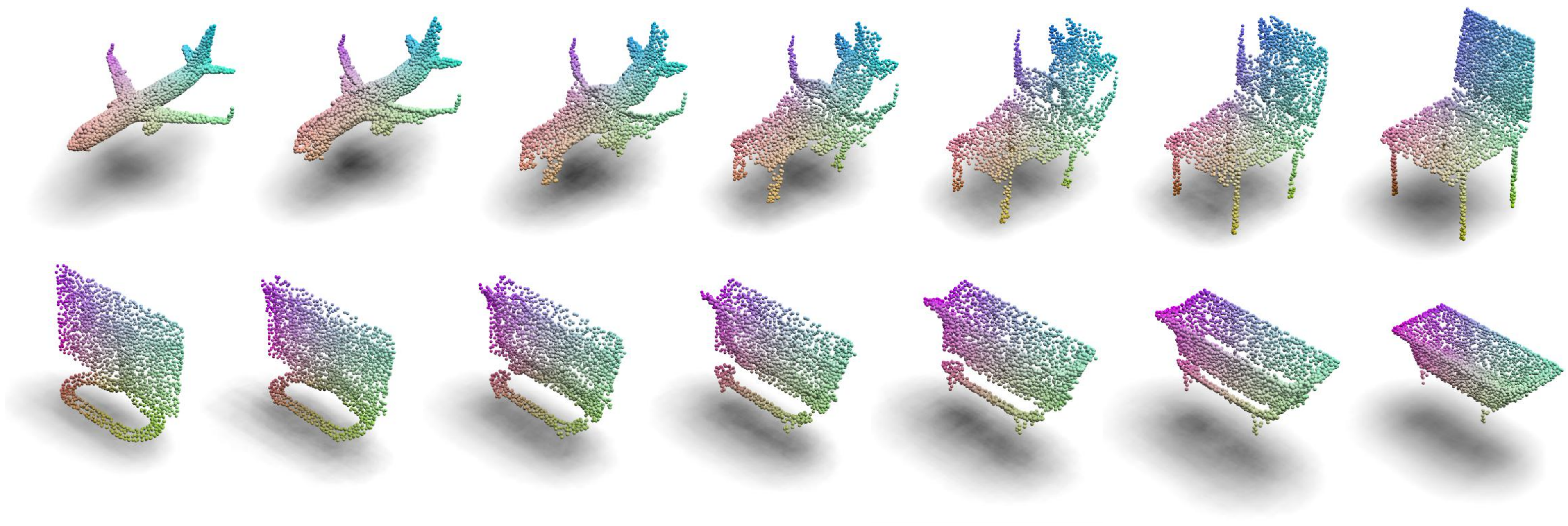}
\end{minipage}
\caption{\textbf{Interpolation between point clouds.} We show the interpolation between examples from different classes (airplane/chair, and monitor/bathtub) with multiple ratios $\lambda$. The interpolants are learned to be classified as $(1-\lambda)$ the first class and $\lambda$ the second class. The interpolation is not obtained by learning, but induced by solving the optimal bijective correspondence which allows the minimum overall distance that each point in one point cloud moves to the assigned point in the other point cloud.}
\label{fig:oam}
\end{figure}

In this work, we make three contributions. First, we introduce data augmentation for point clouds through interpolation and we define the augmentation as a shortest path interpolation. Second, we propose PointMixup, an interpolation between point clouds that computes the optimal assignment as a path function between two point clouds, or the latent representations in terms of point cloud. The proposed interpolation strategy therefore allows usage of successful regularizers of Mixup and Manifold Mixup~\cite{manifoldmixup} on point cloud. We prove that (\textit{i}) our PointMixup indeed finds the shortest path between two point clouds; (\textit{ii}) the assignment does not change for any pairs of the mixed point clouds for any interpolation ratio; and (\textit{iii}) our PointMixup is a linear interpolation, an important property since labels are also linearly interpolated. Figure~\ref{fig:oam} shows two pairs of point clouds, along with our interpolations. Third, we show the empirical benefits of our data augmentation across various tasks, including classification, few-shot learning, and semi-supervised learning. We furthermore show that our approach is agnostic to the network used for classification, while we also become more robust to noise and geometric transformations to the points. 

%% file: 1-relate.tex
\section{Related Work}


\subsubsection{Deep learning for point clouds.}
Point clouds are unordered sets and hence
early works focus on analyzing equivalent symmetric functions which ensures permutation invariance.~\cite{ravanbakhsh2016deep,qi2017pointnet,zaheer2017deep}.
The pioneering PointNet work by Qi~\etal~\cite{qi2017pointnet} presented the first deep network that operates directly on unordered point sets. It learns the global feature with shared multi-layer perceptions and a max pooling operation to ensure permutation invariance. PointNet++~\cite{qi2017pointnet++} extends this idea further with hierarchical structure by relying on a heuristic method of farthest point sampling and grouping to build the hierarchy. Likewise, other recent methods follow to learn hierarchical local features either by grouping points in various manners~\cite{li2018sonet,li2018pointcnn,dgcnn,wu2019pointconv,xu2018spidercnn,zhang2019shellnet,thomas2019kpconv}.
Li \etal~\cite{li2018pointcnn} propose to learn a transformation from the input points to simultaneously solve the weighting of input point features  and  permutation of points into a latent and potentially canonical order. Xu \etal~\cite{xu2018spidercnn} extends 2D convolution
to 3D point clouds by parameterizing a family of convolution filters. Wang \etal~\cite{dgcnn} proposed to leverage neighborhood structures in both point and feature spaces. 

%
%
In this work, we aim to improve point cloud classification for any point-based approach. To that end, we propose a new model-agnostic data augmentation. We propose a Mixup regularization for point clouds and show that it can build on various architectures
to obtain better classification results by reducing the generalization error in classification tasks.
A very recent work by Li \etal~\cite{li2020pointaugment} also considers improving point cloud classification by augmentation. They rely on auto-augmentation and a complicated adversarial training procedure, whereas in this work we propose to augment point clouds by interpolation.

\subsubsection{Interpolation-based regularization.} Employing regularization approaches for training deep neural networks to improve their generalization performances have become standard practice in deep learning. Recent works consider a regularization by interpolating the example and label pairs, commonly known as Mixup~\cite{tokozume2018between,adamixup,mixup}. Manifold Mixup~\cite{manifoldmixup} extends Mixup by interpolating the hidden representations at multiple layers. Recently, an effort has been made on applying Mixup to various tasks such as object detection~\cite{zhang2019bag} and segmentation~\cite{french2019cowmix}.
Different from existing works, which are predominantly employed in the image domain, we propose a new optimal assignment Mixup paradigm for point clouds, in order to deal with their permutation-invariant nature.

Recently, Mixup~\cite{mixup} has also been investigated from a semi-supervised learning perspective\cite{berthelot2019mixmatch,ict,berthelot2019remixmatch}. Mixmatch~\cite{berthelot2019mixmatch} guesses low-entropy labels for unlabelled data-augmented examples and mixes labelled and unlabelled data using Mixup~\cite{mixup}. Interpolation Consistency Training~\cite{ict} utilizes the consistency constraint between the interpolation of unlabelled points with the interpolation of the predictions at those points. 
In this work, we show that our PointMixup can be integrated in such frameworks to enable semi-supervised learning for point clouds.

%% file: 2-method-v3.tex
\section{Point cloud augmentation by interpolation}

\subsection{Problem setting}
In our setting, we are given a training set $\{(S_m,c_m)\}_{m=1}^{M}$ consisting of $M$ point clouds.
$S_m = \{p^{m}_n\}_{n=1}^{N} \in \mathcal{S}$ is a point cloud consisting of $N$ points,  $p^{m}_n \in \mathbb{R}^3$ is the 3D point, $\mathcal{S}$ is the set of such 3D point clouds with $N$ elements. 
$c_m \in \{0,1\}^C$ is the one-hot class label for a total of $C$ classes. The goal is to train a function $h: \mathcal{S} \mapsto [0,1]^C$ that learns to map a point cloud to a semantic label distribution. 
Throughout our work, we remain agnostic to the type of function $h$ used for the mapping and we focus on data augmentation to generate new examples. 

Data augmentation is an integral part of training deep neural networks, especially when the size of the training data is limited compared to the size of the model parameters.
A popular data augmentation strategy is Mixup~\cite{mixup}.
Mixup performs augmentation in the image domain by linearly interpolating pixels, as well as labels.
Specifically, let $I_1 \in \mathbb{R}^{W \times H \times 3}$ and $I_2 \in \mathbb{R}^{W \times H \times 3}$ denote two images.
Then a new image and its label are generated as:
\begin{align}
\label{eq:image_MixUp} I_{\text{mix}} (\lambda) & = (1 - \lambda)\cdot I_1 + \lambda\cdot I_2, \\
c_{\text{mix}} (\lambda) & = (1-\lambda) \cdot c_1 + \lambda \cdot c_2,
\end{align}
where $\lambda \in [0,1]$ denotes the mixup ratio. Usually $\lambda$ is sampled from a beta distribution $\lambda \sim \text{Beta}(\gamma, \gamma)$.
Such a direct interpolation is feasible for images as the data is aligned.
In point clouds, however, linear interpolation is not straightforward.
The reason is that point clouds are sets of points in which the point elements are orderless and permutation-invariant.
We must, therefore, seek a definition of interpolation on unordered sets.



\subsection{Interpolation between point clouds}

Let $S_1 \in \mathcal{S}$ and $S_2 \in  \mathcal{S}$ denote two training examples on which we seek to perform interpolation with ratio $\lambda$ to generate new training examples. 
Given a pair of source examples $S_1$ and $S_2$, an interpolation function, $f_{S_1 \to S_2}: [0,1] \mapsto \mathcal{S}$ 
can be \emph{any continuous function}, which forms a curve that joins $S_1$ and $S_2$ in a metric space $(\mathcal{S}, d)$ with a proper distance function $d$.
This means that it is up to us to define what makes an interpolation good. We define the concept of shortest-path interpolation in the context of point cloud:
\begin{definition}[Shortest-path interpolation]
In a metric space $(\mathcal{S}, d)$, a shortest-path interpolation $f^*_{S_1 \to S_2}: [0,1] \mapsto \mathcal{S}$ is an interpolation between the given pair of source examples $S_1 \in \mathcal{S}$ and $S_2 \in  \mathcal{S}$,  such that for any $\lambda \in [0,1]$, $ d(S_1, S^{(\lambda)}) + d(S^{(\lambda)},S_2)) = d(S_1, S_2)$ holds for $S^{(\lambda)} = f^*_{S_1 \to S_2} (\lambda)$ being the interpolant.
\end{definition}
We say that Definition 1 ensures the shortest path property because the triangle inequality holds for any properly defined distance $d$ : $ d(S_1, S^{(\lambda)}) + d(S^{(\lambda)},S_2)) \geq d(S_1, S_2)$. 
The intuition behind this definition is that the shortest path property ensures the uniqueness of the label distribution on the interpolated data.
To put it otherwise, when computing interpolants from different sources, the interpolants generated by the shortest-path interpolation is more likely to be discriminative than the ones generated by a non-shortest-path interpolation.

\begin{figure}[t!]
\centering
\begin{minipage}[]{0.6\textwidth}
\centering
\includegraphics[width=\textwidth]{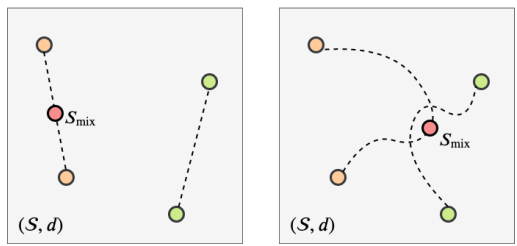}
\end{minipage}
\label{fig:shortest_path}
\caption{\textbf{Intuition of shortest-path interpolation.} The examples lives on a metric space $(\mathcal{S}, d)$ as dots in the figure. The dashed lines are the interpolation paths between different pairs of examples. When the shortest-path property is ensured (left), the interpolation paths from different pairs of source examples are likely to be not intersect in a complicated metric space. While in non-shortest path interpolation (right), the paths can intertwine with each other with a much higher probability, making it hard to tell which pair of source examples does the mixed data come from.}\vspace{-1em}
\end{figure}

%
%
%

To define an interpolation for point clouds, therefore, we must first select a reasonable distance metric.
Then, we opt for the shorterst-path interpolation function based on the selected distance metric.
For point clouds a proper distance metric is the Earth Mover's Distance (EMD), as it captures well not only the geometry between two point clouds, but also local details as well as density distributions~\cite{fan2017point,achlioptas2018learning,liu2019morphing}.
EMD measures the least amount of total displacement required for each of the points in the first point cloud, $x_{i} \in S_1$, to match a corresponding point in the second point cloud, $y_{j} \in S_2$.
Formally, the EMD for point clouds solves the following assignment problem: 
\begin{equation}
    \phi^{*} =  \argmin_{\phi \in \mathbf{\Phi}} \sum_{i} \|x_{i} - y_{\phi(i)}\|_{2},
    \label{eq:optimal_assignment}
\end{equation}
where $ \mathbf{\Phi}=\{ \{1,\dots, N\} \mapsto \{1, \dots, N\} \}$ is the set of possible bijective assignments, which give one-to-one correspondences between points in the two point clouds.
Given the optimal assignment $\phi^*$, the EMD is then defined as the average effort to move $S_1$ points to $S_2$:
\begin{equation}
    d_\text{EMD} = \frac{1}{N} \sum_{i} \|x_{i} - y_{\phi^*(i)}\|_{2}.\vspace{-1em}
\end{equation}
\subsection{PointMixup: Optimal assignment interpolation for point clouds}

We propose an interpolation strategy, which can be used for augmentation that is analogous of Mixup~\cite{mixup} but for point clouds. 
We refer to this proposed PointMixup as \emph{Optimal Assignment (OA) Interpolation}, as it relies on the optimal assignment on the basis of the EMD to define the interpolation between clouds.
Given the source pair of point clouds $S_1 = \{ x_i \}_{i=1}^{N}$ and $S_2 = \{ y_j \}_{j=1}^{N}$, the Optimal Assignment (OA) interpolation is a path function $f^*_{S_1 \to S_2}: [0,1] \mapsto \mathcal{S}$. With $\lambda \in [0,1]$,
%
\begin{align}
    f^*_{S_1 \to S_2} (\lambda)&= \{ u_i \}_{i=1}^{N},  \quad \text{where}\\
    u_i = (1-\lambda) &\cdot x_i + \lambda \cdot y_{\phi^*(i)},  \label{eq:oaMixUp}
\end{align}
in which $\phi^*$ is the optimal assignment from $S_1$ to $S_2$ defined by Eq.~\ref{eq:optimal_assignment}.
%
%
%
Then the interpolant $S_\textbf{OA}^{S_1 \to S_2, (\lambda)}$ (or $S_\textbf{OA}^{(\lambda)} $ when there is no confusion) generated by the OA interpolation path function $f^*_{S_1 \to S_2} (\lambda)$ is the required augmented data for point cloud Mixup.
\begin{equation}
    S_\textbf{OA}^{(\lambda)} = \{ (1-\lambda) \cdot x_i + \lambda \cdot y_{\phi^*(i)}\}_{i=1}^{N}.
\end{equation}
Under the view of $f^*_{S_1 \to S_2}$ being a path function in the metric space $(\mathcal{S}, d_\text{EMD})$, $f$ is expected to be the shortest path joining $S_1$ and $S_2$ since the definition of the interpolation is induced from the EMD.




\subsection{Analysis}




Intuitively we expect that PointMixup is a shortest path linear interpolation.
That is, the interpolation lies on the shortest path joining the source pairs, and the interpolation is linear with regard to $\lambda$ in  $(\mathcal{S}, d_\text{EMD})$, since the definition of the interpolation is derived from the EMD.
However, it is non-trivial to show the optimal assignment 
interpolation abides to a shortest path linear interpolation, because the optimal assignment between the mixed point cloud and either of the source point cloud is unknown.
It is, therefore, not obvious that we can ensure whether there exists a shorter path between the mixed examples and the source examples.
To this end, we need to provide an in-depth analysis. 

To ensure the uniqueness of the label distribution from the mixed data, we need to show that the \emph{shortest path property} w.r.t. the EMD is fulfilled. 
Moreover, we need to show that the proposed interpolation is \emph{linear} w.r.t the EMD, in order to ensure that the input interpolation has the same ratio as the label interpolation. Besides, we evaluate the \emph{assignment invariance property} as a prerequisite knowledge for the proof for the linearity.
This property implies that there exists no shorter path between interpolants with different $\lambda$, \ie, the shortest path between the interpolants is a part of the shortest path between the source examples. Due to space limitation, we sketch the proof for each property. The complete proofs are available in the supplementary material.

We start with the shortest path property. Since the EMD for point cloud is a metric, the triangle inequality $d_{EMD}(A, B)+ d_{EMD}(B,C) \geq d_{EMD} (A,C)$ holds (for which a formal proof can be found in \cite{rubner2000earth}). Thus we formalize the shortest path property into the following proposition:




\begin{property_}[shortest path]
    Given the source examples $S_1$ and $S_2$, $\forall \lambda\in [0,1],$ $d_\text{EMD}(S_1, S_{\textbf{OA}}^{(\lambda)}) + d_\text{EMD}(S_{\textbf{OA}}^{(\lambda)}, S_2 ) = d_\text{EMD}(S_1, S_2)$.
\end{property_}

\begin{proof-sketch}
From the definition of the EMD we can derive $d_{\text{EMD}} (S_1, S_{\textbf{OA}}^{(\lambda)}) + d_{\text{EMD}} (S_2, S_{\textbf{OA}}^{(\lambda)}) \leq d_{\text{EMD}} (S_1, S_2)$. Then from the triangle inequity of the EMD, only the equality remains.
\end{proof-sketch}

We then introduce the assignment invariance property of the OA Mixup as an intermediate step for the proof of the linearity of OA Mixup. The property shows that the assignment does not change for any pairs of the mixed point clouds with different $\lambda$. Moreover, the assignment invariance property is important to imply that the shortest path between the any two mixed point clouds is part of the shortest path between the two source point clouds.

\begin{property_}[assignment invariance]
$S_{\textbf{OA}}^{(\lambda_1)}$ and $S_{\textbf{OA}}^{(\lambda_2)}$ are two mixed point clouds from the same given source pair of examples $S_1$ and $S_2$ as well as the mix ratios $\lambda_1$ and $\lambda_2$ such that $0\leq \lambda_1 < \lambda_2 \leq 1$. Let the points in $S_{\textbf{OA}}^{(\lambda_1)}$ and $S_{\textbf{OA}}^{(\lambda_2)}$ be $u_i =  (1-\lambda_1) \cdot x_i + \lambda_1 \cdot y_{\phi^*(i)}$ and $v_k =  (1-\lambda_2) \cdot x_k + \lambda_2 \cdot y_{\phi^*(k)}$, where $\phi^*$ is the optimal assignment from $S_1$ to $S_2$. Then the identical assignment $\phi_I$ is the optimal assignment from $S_{\textbf{OA}}^{(\lambda_1)}$ to $S_{\textbf{OA}}^{(\lambda_2)}$. 
\end{property_}

\begin{proof-sketch}
We first prove that the identical mapping is the optimal assignment from  $S_1$ to $S_{\textbf{OA}}^{(\lambda_1)}$ from the definition of the EMD. Then we prove that  $\phi^*$ is the optimal assignment from  $S_{\textbf{OA}}^{(\lambda_1)}$ to $S_2$. Finally we prove that the identical mapping is the optimal assignment from $S_{\textbf{OA}}^{(\lambda_1)}$ to $S_{\textbf{OA}}^{(\lambda_2)}$ similarly as the proof for the first intermediate argument.
\end{proof-sketch}

\noindent Given the property of  assignment invariance, the linearity follows:

\begin{property_}[linearity]
For any mix ratios $\lambda_1$ and $\lambda_2$ such that $0\leq \lambda_1 < \lambda_2 \leq 1$, the mixed point clouds $S_{\textbf{OA}}^{(\lambda_1)}$ and $S_{\textbf{OA}}^{(\lambda_2)}$ satisfies that $d_\text{EMD}(S_{\textbf{OA}}^{(\lambda_1)}, S_{\textbf{OA}}^{(\lambda_2)}) = (\lambda_2 -\lambda_1) \cdot d_\text{EMD}(S_1, S_2)$.
\end{property_}

\begin{proof-sketch}
The proof can be directly derived from the fact that the identical mapping is the optimal assignment between $S_{\textbf{OA}}^{(\lambda_1)}$ and $S_{\textbf{OA}}^{(\lambda_2)}$.
\end{proof-sketch}

The linear property of our interpolation is important, as we jointly interpolate the point clouds and the labels. By ensuring that the point cloud interpolation is linear, we ensure that the input interpolation has the same ratio as the label interpolation. 

On the basis of the properties, we find that PointMixup is a shortest path linear interpolation between point clouds in $(\mathcal{S}, d_\text{EMD})$. 

\subsection{Manifold PointMixup: Interpolate between latent point features}

In standard PointMixup, only the inputs, \ie, the XYZ point cloud coordinates are mixed. The input XYZs are low-level geometry information and sensitive to disturbances and transformations, which in turn limits the robustness of the PointMixup. Inspired by Manifold Mixup~\cite{manifoldmixup}, we can also use the proposed interpolation solution to mix the latent representations in the hidden layers of point cloud networks,
which are trained to capture salient and high-level information that is less sensitive to transformations. PointMixup can be applied for the purpose of Manifold Mixup to mix both at the XYZs and different levels of latent point cloud features and maintain their respective advantages, which is expected to be a stronger regularizer for improved performance and robustness.


We describe how to mix the latent representations. Following \cite{manifoldmixup}, at each batch we randomly select a layer $l$ to perform PointMixup from a set of layers $L$, which includes the input layer.
In a point cloud network network, the intermediate latent representation at layer $l$ (before the global aggregation stage such as the max pooling aggregation in PointNet~\cite{qi2017pointnet} and PointNet++~\cite{qi2017pointnet++}) is
$Z_{(l)} = \{(x_i, z_i^{(x)})\}_{i=1}^{N_z}$, in which $x_i$ is 3D point coordinate and $z_i^{(x)}$ is the corresponding high-dimensional feature. For the mixed latent representation, given the latent representation of two source examples are $Z_{(l),1} = \{(x_i, z_i^{(x)})\}_{i=1}^{N_z}$ and $Z_{(l),2} = \{(y_i, z_i^{(y)})\}_{i=1}^{N_z}$, the optimal assignment $\phi^*$ is obtained by the 3D point coordinates $x_i$, and the mixed latent representation then becomes 
\begin{align*}
    Z_{(l),\textbf{OA}}^{(\lambda)} &= \{(x^{\text{mix}}_i, z_i^{\text{mix}})\}, \quad \quad \text{where} \\
    x^{\text{mix}}_i &= (1-\lambda) \cdot x_i + \lambda \cdot y_{\phi^*(i)}, \\
    z_i^{\text{mix}} &=  (1-\lambda) \cdot z_i^{(x)} + \lambda \cdot z^{(y)}_{\phi^*(i)}.
\end{align*}
Specifically in PointNet++,  three layers of representations are randomly selected to perform Manifold Mixup: the input, and the representations after the first and the second SA modules (See appendix of \cite{qi2017pointnet++}). 


%% file: 3-exp-v3.tex
\section{Experiments}

\subsection{Setup}

\textbf{Datasets.}
We focus in our experiments on the \textbf{ModelNet40} dataset~\cite{modelnet40}. This dataset contains 12,311 CAD models
from 40 man-made object categories, split into 9,843 for training and 2,468 for testing. We furthermore perform experiments on the \textbf{ScanObjectNN} dataset~\cite{uy2019scanobjnn}. This dataset consists of real-world point cloud objects, rather than sampled virtual point clouds. The dataset consists of 2,902 objects and 15 categories. We report on two variants of the dataset, a standard variant OBJ\_ONLY and one with heavy permutations from rigid transformations PB\_T50\_RS~\cite{uy2019scanobjnn}.

Following \cite{li2018pointcnn}, we discriminate between settings where each dataset is pre-aligned and unaligned with horizontal rotation on training and test point cloud examples. For the unaligned settings, we randomly rotate the training point cloud along the up-axis.
Then, before solving the optimal assignment, we perform a simple additional alignment step to fit and align the symmetry axes between the two point clouds to be mixed.
Through this way, the point clouds are better aligned and we obtain more reasonable point correspondences. 
Last, we also perform experiments using only 20\% of the training data.

\textbf{Network architectures.}
The main network architecture used throughout the paper is PointNet++~\cite{qi2017pointnet++}.  We also report results with PointNet~\cite{qi2017pointnet} and DGCNN~\cite{dgcnn}, to show that our approach is agnostic to the architecture that is employed. PointNet learns a permutation-invariant set function, which does not capture local structures induced by the metric space the points live in. PointNet++ is a hierarchical structure, which segments a point cloud into smaller clusters and applies PointNet locally. DGCNN performs hierarchical operations by selecting a local neighbor in the feature space instead of the point space, resulting in each point having different neighborhoods in different layers.

\textbf{Experimental details.}
We uniformly sample 1,024 points on the mesh faces according to the face area and normalize them to be contained in a unit sphere, which is a standard setting~\cite{qi2017pointnet,qi2017pointnet++,li2018pointcnn}.
In case of mixing clouds with different number of points, we can simply replicate random elements from the each point set to reach the same cardinality. During training, we augment the point clouds on-the-fly with random jitter for each point using Gaussian noise with zero mean and 0.02 standard deviation. We implement our approach in PyTorch~\cite{pytorch}. For network optimization, we use the Adam optimizer with an initial learning rate of $10^{-3}$. The model is trained for 300 epochs with a batch size of 16. We follow previous work~\cite{mixup,manifoldmixup} and draw $\lambda$ from a beta distribution $\lambda \sim \text{Beta} (\gamma, \gamma)$. We also perform Manifold Mixup~\cite{manifoldmixup} in our approach, through interpolation on the transformed and pooled points in intermediate network layers. In this work, we opt to use the efficient algorithm and adapt the open-source implementation from \cite{liu2019morphing} to solve the optimal assignment approximation.
Training for 300 epochs takes around 17 hours without augmentation and around 19 hours with PointMixup or Manifold PointMixup on a single NVIDIA GTX~1080~ti.

\begin{figure}[t]
\centering
\begin{minipage}[]{1.0\textwidth}
\centering
\includegraphics[width=\textwidth]{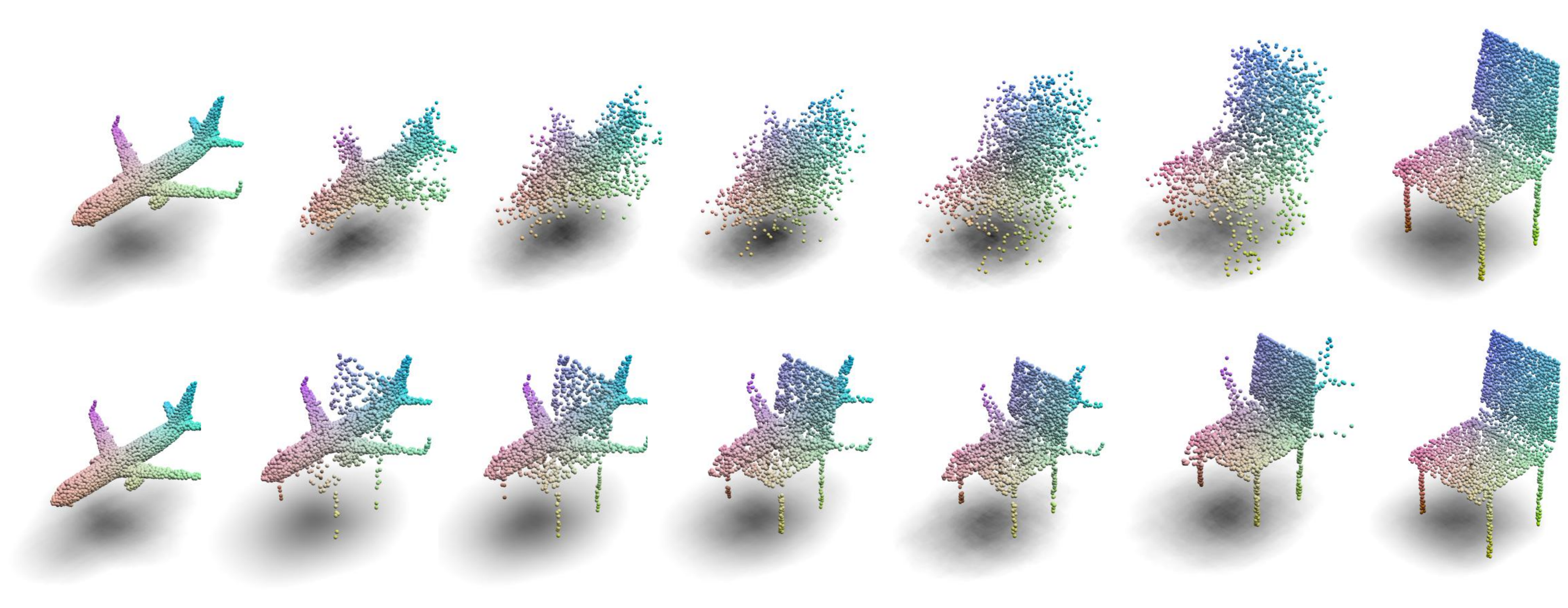}
\label{fig:psra}
\end{minipage}
\caption{\textbf{Baseline interpolation variants.} Top: point cloud interpolation through random assignment. Bottom: interpolation through sampling.}\vspace{-1em}
\end{figure}

\textbf{Baseline interpolations.}
For our comparisons to baseline point cloud augmentations, we compare to two variants. The first variant is random assignment interpolation, where a random assignment $\phi^\textbf{RA}$ is used, to connect points from both sets, yielding: $$S_{\textbf{RA}}^{(\lambda)} = \{(1-\lambda) \cdot x_i + \lambda \cdot y_{\phi^\textbf{RA}(i)}\}.$$
The second variant is point sampling interpolation, where random draws without replacement of points from each set are made according to the sampling frequency $\lambda$:
    $$S_{\textbf{PS}}^{(\lambda)} = S_1^{(1-\lambda)} \cup S_2^{(\lambda)},$$
    where $S_2^{(\lambda)}$ denotes a randomly sampled subset of $S_2$, with $ \lfloor \lambda N \rfloor$ elements. ($\lfloor \cdot \rfloor$ is the floor function.) And similar for $S_1$ with $N - \lfloor \lambda N \rfloor$ elements, such that $S_{\textbf{PS}}^{(\lambda)}$ contains exactly $N$ points.
The intuition of the point sampling variant is that for point clouds as unordered sets, one can move one point cloud to another through a set operation such that it removes several random elements from set $S_1$ and replace them with same amount of elements from $S_2$.

\subsection{Point cloud classification ablations}
We perform four ablation studies to show the workings of our approach with respect to the interpolation ratio, comparison to baseline interpolations and other regularizations, as well robustness to noise.

\begin{wrapfigure}{r}{55mm}
\centering
\vspace{-2em}
\begin{minipage}[t]{1\linewidth}
\centering
\includegraphics[width=1\textwidth]{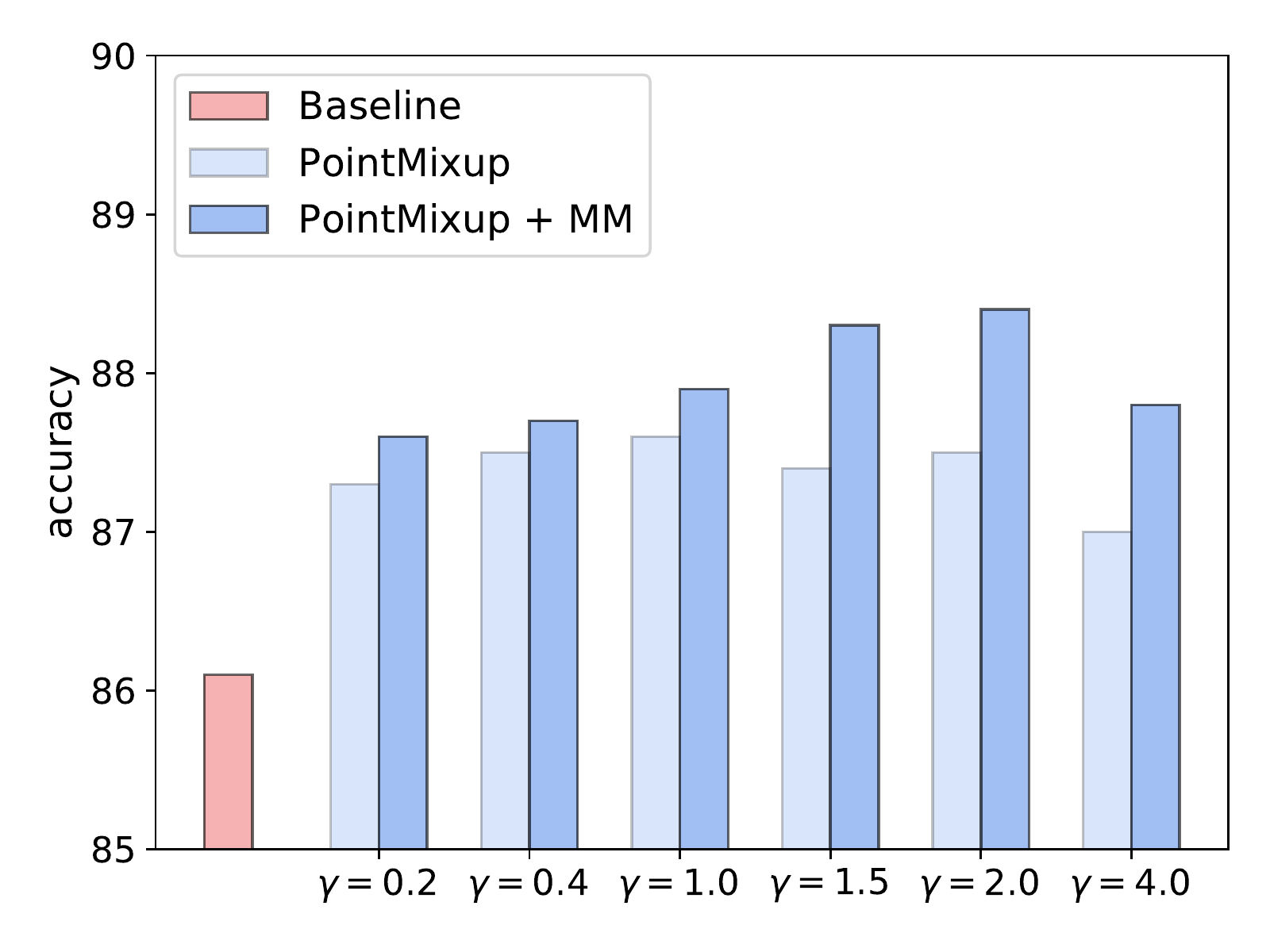}
\end{minipage}
\vspace{-1em}
\caption{\textbf{Effect of interpolation ratios.} MM denotes Manifold Mixup.}
\vspace{-1.5em}
\label{fig:ablation1}
\end{wrapfigure} 
\textbf{Effect of interpolation ratio.} The first ablation study focuses on the effect of the interpolation ratio in the data augmentation for point cloud classification. We perform this study on ModelNet40 using the PointNet++ architecture. The results are shown in Fig.~\ref{fig:ablation1} for the pre-aligned setting. We find that regardless of the interpolation ratio used, our approach provides a boost over the setting without augmentation by interpolation. PointMixup positively influences point cloud classification. The inclusion of manifold mixup adds a further boost to the scores. Throughout further experiments, we use $\gamma=0.4$ for input mixup and $\gamma=1.5$ for manifold mixup in unaligned setting, and $\gamma=1.0$ for input mixup and $\gamma=2.0$ for manifold mixup in pre-aligned setting.

\textbf{Comparison to baseline interpolations.}
In the second ablation study, we investigate the effectiveness of our PointMixup compared to the two interpolation baselines. We again use ModelNet40 and PointNet++. We perform the evaluation on both the pre-aligned and unaligned dataset variants, where for both we also report results with a reduced training set.
The results are shown in Table~\ref{tab:main}. Across both the alignment variants and dataset sizes, our PointMixup obtains favorable results. This result highlights the effectiveness of our approach, which abides to the shortest path linear interpolation definition, while the baselines do not.

\begin{table}[t]
\centering
\caption{\textbf{Comparison of PointMixup to baseline interpolations} on ModelNet40 using PointNet++. PointMixup compares favorable to excluding interpolation and to the baselines, highlighting the benefits of our shortest path interpolation solution.}
\resizebox{0.99\columnwidth}{!}{%
\setlength\tabcolsep{4pt} 
\begin{tabular}{lccccccc} 
\toprule
 & \textbf{No mixup} & \multicolumn{2}{c}{\textbf{Random assignment}} & \multicolumn{2}{c}{\textbf{Point sampling}} & \multicolumn{2}{c}{\textbf{PointMixup} }\\
 \cmidrule(lr){2-2} \cmidrule(lr){3-4} \cmidrule(lr){5-6} \cmidrule(lr){7-8}
 Manifold mixup & $\times$ & $\times$ & $\checkmark$ & $\times$ & $\checkmark$ & $\times$ & $\checkmark$\\
\midrule
\rowcolor{mygray}
\textbf{Full dataset} & & & & & & &\\
Unaligned & 90.7 & 90.8 & 91.1 & 90.9 & {91.4} & 91.3 & \textbf{91.7}\\
Pre-aligned & 91.9 & 91.6 & 91.9 & 92.2 & 92.5 & 92.3 & \textbf{92.7}\\
\hline
\rowcolor{mygray}
\textbf{Reduced dataset} & & & & & & &\\
Unaligned & 84.4 & 84.8 & 85.4 & 85.7 & 86.5 & 86.1 & \textbf{86.6}\\
Pre-aligned & 86.1 & 85.5 & 87.3 & 87.2 & 87.6 & 87.6 & \textbf{88.6}\\
\bottomrule
\end{tabular}
}
\label{tab:main}
\end{table}
\begin{table}[t!]
\vspace{-1em}
\caption{\textbf{Evaluating our approach to other data augmentations} (left) \textbf{and its robustness to noise and transformations} (right). We find that our approach with manifold mixup (MM) outperforms augmentations such as label smoothing and other variations of mixup. For the robustness evaluation, we find that our approach with strong regularization power from manifold mixup provides more robustness to random noise and geometric transformations.}
\label{tab:regu}
\centering
\begin{subtable}{0.48\linewidth}
\centering
\adjustbox{height=2.4cm}{
\begin{tabular}{lcc}
\toprule
 & \multicolumn{2}{c}{\textbf{PointMixup}}\\ \cmidrule(lr){2-3}
 & $\times$ & $\checkmark$\\
\midrule
Baseline with no mixing & 86.1 & -- \\
Mixup &-- & 87.6\\
Manifold mixup &-- &\textbf{88.6}\\
\midrule
Mix input, not labels &-- & 86.6\\
Mix input from same class &-- & 86.4\\ 
\midrule
Mixup latent (layer 1) & -- & 86.9\\
Mixup latent (layer 2) & -- & 86.8\\
\midrule
Label smoothing (0.1) & 87.2 & -- \\
Label smoothing (0.2) & 87.3 & -- \\
\bottomrule
\end{tabular}
}
\end{subtable}
~
\begin{subtable}{0.48\linewidth}
\centering
\adjustbox{height=2.4cm}{
\begin{tabular}{lcccc}
\toprule
\textbf{Transforms} & \multicolumn{3}{c}{\textbf{PointMixup}}\\ \cmidrule(lr){2-4}
 & $\times$ & w/o MM & w/ MM\\ 
\midrule
Noise $\sigma^{2}=0.01$&91.3&91.9&\textbf{92.3}\\
Noise $\sigma^{2}=0.05$&35.1&51.5&\textbf{56.5}\\
Noise $\sigma^{2}=0.1$&4.03&4.27&\textbf{7.42}\\
\midrule
Z-rotation [-30,30] &74.3&70.9&\textbf{77.8}\\
X-rotation [-30,30]&73.2&70.8&\textbf{76.8}\\
Y-rotation [-30,30]&87.6&87.9&\textbf{88.7}\\
\midrule
Scale (0.6)&85.8&84.5&\textbf{86.3}\\
Scale (2.0)&59.2&67.7&\textbf{72.9}\\
\midrule
DropPoint (0.2)&84.9&78.1&\textbf{90.9}\\
\bottomrule
\end{tabular}
}
\end{subtable}
\end{table}

\textbf{PointMixup with other regularizers.}
Third, we evaluate how well PointMixup works by comparing to multiple existing data regularizers and mixup variants, again on ModelNet40 and PointNet++. We investigate the following augmentations: (\textit{i}) Mixup~\cite{mixup}, (\textit{ii}) Manifold Mixup~\cite{manifoldmixup}, (\textit{iii}) mix input only without target mixup, (\textit{iv}) mix latent representation at a fixed layer (manifold mixup does so at random layers), and (\textit{v}) label smoothing~\cite{szegedy2016rethinking}. Training is performed on the reduced dataset to better highlight their differences.
We show the results in Table~\ref{tab:regu} on the left. Our approach with manifold mixup obtains the highest scores. The label smoothing regularizer is outperformed, while we also obtain better scores than the mixup variants. We conclude that PointMixup is forms an effective data augmentation for point clouds.

\textbf{Robustness to noise.}
By adding additional augmented training examples, we enrich the dataset. This enrichment comes with additional robustness with respect to noise in the point clouds. We evaluate the robustness by adding random noise perturbations on point location, scale, translation and different rotations. Note that for evaluation of robustness against up-axis rotation, we use the models which are trained with the pre-aligned setting, in order to test also the performance against rotation along the up-axis as a novel transform. The results are in Table~\ref{tab:regu} on the right.
Overall, our approach including manifold mixup provides more stability across all perturbations. For example, with additional noise ($\sigma=0.05$), we obtain an accuracy of 56.5, compared to 35.1 for the baseline. We similar trends for scaling (with a factor of two), with an accuracy of 72.9 versus 59.2. We conclude that PointMixup makes point cloud networks such as PointNet++ more stable to noise and rigid transformations.

\begin{figure}[t]
\centering
\begin{minipage}[]{0.47\textwidth}
\centering
\includegraphics[width=\textwidth]{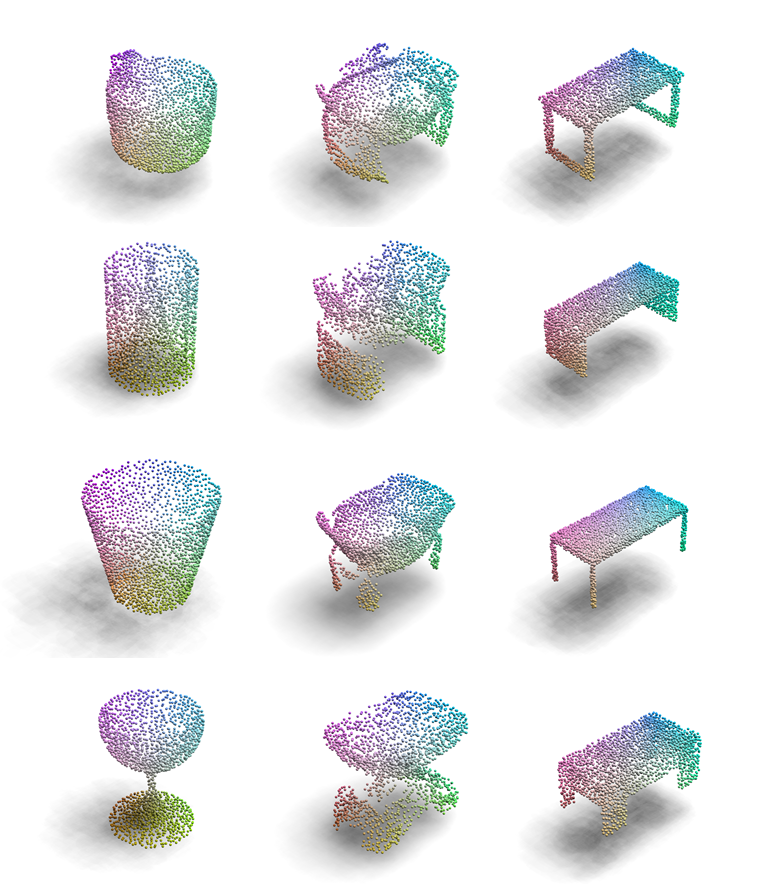}
\end{minipage}
\begin{minipage}[]{0.47\textwidth}
\centering
\includegraphics[width=\textwidth]{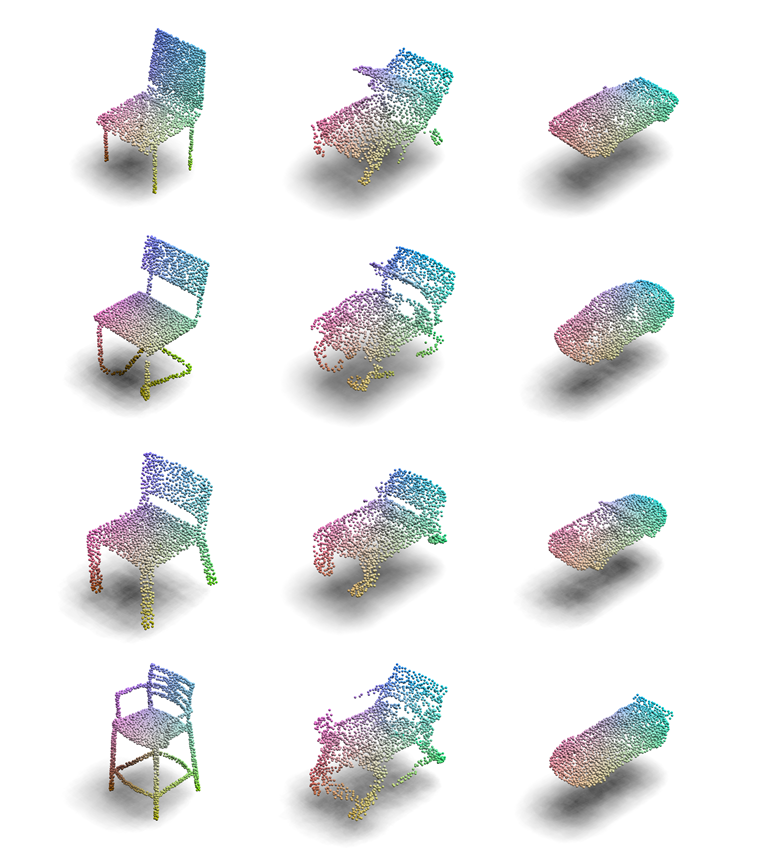}
\end{minipage}
\caption{\textbf{Qualitative examples of PointMixup.} We provide eight visualizations of our interpolation. The four examples on the left show interpolations for different configurations of cups and tables. The four examples on the right show interpolations for different chairs and cars.}\vspace{-1em}
\label{fig:qual}
\end{figure}

\textbf{Qualitative analysis.}
In Figure~\ref{fig:qual}, we show eight examples of PointMix for point cloud interpolation; four interpolations of cups and tables, four interpolations of chairs and cars. Through our shortest path interpolation, we end up at new training examples that exhibit characteristics of both classes, making for sensible point clouds and mixed labels, which in turn indicate why PointMixup is beneficial for point cloud classification.

\subsection{Evaluation on other networks and datasets}
With PointMixup, new point clouds are generated by interpolating existing point clouds. As such, we are agnostic to the type of network or dataset. To highlight this ability, we perform additional experiments on extra networks and an additional point cloud dataset.

\begin{table}[t]
\caption{\textbf{PointMixup on other networks} (left) \textbf{and another dataset} (right). We find our approach is beneficial regardless the network or dataset.}
\label{tab:other}
\centering
\begin{subtable}{0.5\linewidth}\centering
\adjustbox{height=0.9cm}{
\setlength\tabcolsep{3pt} 
\begin{tabular}{lccccc}
\toprule
&\multicolumn{2}{c}{\textbf{PointNet}} &\multicolumn{3}{c}{\textbf{DGCNN}} \\ \cmidrule(lr){2-3} \cmidrule(lr){4-6}
& $\times$ & w/o MM & $\times$ & w/o MM & w/ MM\\
\hline
Full & 89.2 & \textbf{89.9} & 92.7 & 92.9 & \textbf{93.1} 
 \\
Reduced & 81.3 & \textbf{83.4} & 88.2 & 88.8 & \textbf{89.0} \\
\bottomrule
\end{tabular}
}
\end{subtable}
~
\begin{subtable}{0.4\linewidth}\centering
\adjustbox{height=0.9cm}{
\setlength\tabcolsep{3pt} 
\begin{tabular}{lccc}
\toprule
 & \multicolumn{3}{c}{\textbf{ScanObjectNN}}\\ \cmidrule(lr){2-4}
 & $\times$ & w/o MM & w/ MM\\
 \hline
Standard & 86.6 & 87.6 & \textbf{88.5} \\
Perturbed & 79.3 & 80.2 & \textbf{80.6} \\ \bottomrule
\end{tabular}
}
\end{subtable}
\end{table}

\textbf{PointMixup on other network architectures.}
We show the effect of PointMixup to two other networks, namely PointNet~\cite{qi2017pointnet} and DGCNN~\cite{dgcnn}. The experiments are performed on ModelNet40. For PointNet, we perform the evaluation on the unaligned setting and for DGCNN with pre-aligned setting to remain consistent with the alignment choices made in the respective papers. The results are shown in Table~\ref{tab:other} on the left. 
We find improvements when including PointMixup for both network architectures.

\textbf{PointMixup on real-world point clouds.}
We also investigate PointMixup on point clouds from real-world object scans, using ScanObjectNN~\cite{uy2019scanobjnn}, which collects object from 3D scenes in SceneNN~\cite{hua2016scenenn} and ScanNet~\cite{dai2017scannet}. Here, we rely on PointNet++ as network. The results in Table~\ref{tab:other} on the right show that we can adequately deal with real-world point cloud scans, hence we are not restricted to point clouds from virtual scans. This result is in line with experiments on point cloud perturbations.

\subsection{Beyond standard classification}
The fewer training examples available, the stronger the need for additional examples through augmentation. 
Hence, we train PointNet++ on ModelNet40 in both a few-shot and semi-supervised setting.

\textbf{Semi-supervised learning.}
Semi-supervised learning learns from a dataset where only a small portion of data is labeled. 
%
Here, we show how PointMixup directly enables semi-supervised learning for point clouds. We start from Interpolation Consistency Training~\cite{ict}, a state-of-the-art semi-supervised approach, which utilizes Mixup between unlabeled points. Here, we use our Mixup for point clouds within their semi-supervised approach. 
We evaluate on ModelNet40 using 400, 600, and 800 labeled point clouds. The result of semi-supervised learning are illustrated in Table~\ref{tab:few} on the left. Compared to the supervised baseline, which only uses the available labelled examples, our mixup enables the use of additional unlabelled training examples, resulting in a clear boost in scores. With 800 labelled examples, the accuracy increases from 73.5\% to 82.0\%, highlighting the effectiveness of PointMixup in a semi-supervised setting.

\begin{table}[t]
\caption{\textbf{Evaluating PointMixup in the context of semi-supervised} (left) \textbf{and few-shot learning} (right). When examples are scarce, as is the case for both settings, using our approach provides a boost to the scores.}
\label{tab:few}
\centering
\begin{subtable}{0.50\linewidth}\centering
\setlength\tabcolsep{4pt} 
\adjustbox{height=1cm}{
\begin{tabular}{lcc}
\toprule
& \multicolumn{2}{c}{\textbf{Semi-supervised classification}}\\ \cmidrule(lr){2-3}
&Supervised&\cite{ict}+PointMixup\\
\hline
400 examples&69.4& \textbf{76.7}\\
600 examples&72.6&\textbf{80.8}\\
800 examples &73.5&\textbf{82.0}\\
\bottomrule
\end{tabular}
}
\end{subtable}
~
\begin{subtable}{0.46\linewidth}\centering
\adjustbox{height=1cm}{
\setlength\tabcolsep{4pt} 
\begin{tabular}{lccc}
\toprule
& \multicolumn{2}{c}{\textbf{Few-shot classification}}\\\cmidrule(lr){2-3}
& \cite{snell2017protonet} &  + PointMixup\\ 
\hline
5-way 1-shot&72.3&\textbf{77.2} \\ 
5-way 3-shot&80.2&\textbf{82.2}\\
5-way 5-shot&84.2&\textbf{85.9}\\
\bottomrule
\end{tabular}
}
\end{subtable}
\end{table}

\textbf{Few-shot learning.}
Few-shot classification aims to learn a classifier to recognize unseen classes during training with limited examples. We follow ~\cite{vinyals2016matching,ravi2016optimization,snell2017protonet,finn2017maml,sung2018learning} to regard few-shot learning a typical meta-learning method, which learns how to learn from limited labeled data through training from a collection of tasks, \ie, episodes. In an $N$-way $K$-shot setting, in each task, $N$ classes are selected and $K$ examples for each class are given as a support set, and the query set consists of the examples to be predicted. 
We perform few-shot classification on ModelNet40, from which we select 20 classes for training, 10 for validation, and 10 for testing. We utilize PointMixup within ProtoNet~\cite{snell2017protonet} by constructing mixed examples from the support set and update the model with the mixed examples before making predictions on the query set. We refer to the supplementary material for the details of our method and the settings. The results in Table~\ref{tab:few} on the right show that incorporating our data augmentation provides a boost in scores, especially in the one-shot setting, where the accuracy increases from 72.3\% to 77.2\%.



%% file: 4-conclude.tex
\section{Conclusion}
This work proposes PointMixup for data augmentation on point clouds. Given the lack of data augmentation by interpolation on point clouds, we start by defining it as a shortest path linear interpolation. We show how to obtain PointMixup between two point clouds by means of an optimal assignment interpolation between their point sets. As such, we arrive at a Mixup for point clouds, or latent point cloud representations in the sense of Manifold Mixup, that can handle permutation invariant nature. We first prove that PointMixup abides to our shortest path linear interpolation definition. Then, we show through various experiments that PointMixup matters for point cloud classification. We show that our approach outperforms baseline interpolations and regularizers.  Moreover, we highlight increased robustness to noise and geometric transformations, as well as its general applicability to point-based networks and datasets. Lastly, we show the potential of our approach in both semi-supervised and few-shot settings. The generic nature of PointMixup allows for a comprehensive embedding in point cloud classification.

\vspace{10pt}
{
\noindent \textbf{Acknowledgment}
This research was supported in part by the SAVI/MediFor and the NWO VENI What \& Where projects. We thank the anonymous reviewers for helpful comments and suggestions.
}

%% file: 6-supp.tex
\section{Proofs for the properties of PointMixup interpolation}

We provide detailed proofs for the shortest path property, the assignment invariance property and the linearity, stated in Section 3.4.
\vspace{1.5em}

\begin{proof1}
We denote $x_i \in S_1$ and $y_j \in S_2$ are the points in $S_1$ and $S_2$, then the generated $S_{\textbf{OA}}^{(\lambda)} = \{u_i\}_{i=1}^N$ and $u_i =  (1-\lambda) \cdot x_i + \lambda \cdot y_{\phi^*(i)}$, where $\phi^*$ is the optimal assignment from $S_1$ to $S_2$.

Then we suppose an identical one-to-one mapping $\phi_I$ such that $\phi_I(i) = i$. Then by definition of the EMD as the minimum 
transportation distance, so
\begin{equation}
    d_{\text{EMD}} (S_1, S_{\textbf{OA}}^{(\lambda)}) \leq \frac{1}{N} \sum_{i} \| x_i - u_{\phi_I(i)} \|_2, \label{eq:phiI}
\end{equation}
where the right term of (\ref{eq:phiI}) is the transportation distance under identical assignment $\phi_I$. Since $ \frac{1}{N} \sum_{i} \| x_i - u_{\phi_I(i)} \|_2 = \frac{1}{N} \sum_{i} \| x_i - ((1-\lambda) \cdot x_i + \lambda \cdot y_{\phi^*(i)})  \|_2 = \lambda\frac{1}{N} \sum_{i} \| x_i -  y_{\phi^*(i)} \|_2 = \lambda \cdot d_{\text{EMD}} (S_1, S_2)$. Thus, 
\begin{align}
\label{eq:ineq1} d_{\text{EMD}} (S_1, S_{\textbf{OA}}^{(\lambda)}) & \leq \lambda \cdot  d_{\text{EMD}} (S_1, S_2).
\end{align}
Similarly as in (\ref{eq:phiI}) and (\ref{eq:ineq1}), the following inequality (\ref{eq:ineq2}) can be derived by assigning  the correspondence from  $S_{\textbf{OA}}^{(\lambda)}$ to $S_2$ with $\phi^*$:
\begin{align}
    d_{\text{EMD}} ( S_{\textbf{OA}}^{(\lambda)}, S_2) & \leq (1-\lambda) \cdot  d_{\text{EMD}} (S_1, S_2)\label{eq:ineq2}.
\end{align}
With (\ref{eq:ineq1}) and (\ref{eq:ineq2}),
 \begin{equation}
      d_{\text{EMD}} (S_1, S_{\textbf{OA}}^{(\lambda)}) + d_{\text{EMD}} (S_2, S_{\textbf{OA}}^{(\lambda)}) \leq d_{\text{EMD}} (S_1, S_2).\label{eq:ineq_sum}
 \end{equation}
However, as the triangle inequality holds for the EMD, \ie
 \begin{equation}
      d_{\text{EMD}} (S_1, S_{\textbf{OA}}^{(\lambda)}) + d_{\text{EMD}} (S_2, S_{\textbf{OA}}^{(\lambda)}) \geq d_{\text{EMD}} (S_1, S_2),\label{eq:triangle}
 \end{equation}
 Then by summarizing (\ref{eq:ineq_sum}) and (\ref{eq:triangle}), $d_{\text{EMD}} (S_1, S_{\textbf{OA}}^{(\lambda)}) + d_{\text{EMD}} (S_2, S_{\textbf{OA}}^{(\lambda)})\\ = d_{\text{EMD}} (S_1, S_2)$ is proved.
\end{proof1}

\begin{proof2}
We introduce two intermediate arguments.
We begin with proving the first intermediate argument: $\phi_I$ is the optimal assignment from $S_1$ to $S_{\textbf{OA}}^{(\lambda_1)}$. Similarly as in (\ref{eq:ineq1}) ,(\ref{eq:ineq2}) and (\ref{eq:triangle}) from the proof for Proposition 1, in order to allow all the three inequalities, the equal signs need to be taken for all of the three inequalities. Consider that the equal sign being taken for (\ref{eq:ineq1}) is equivalent to the the equal sign being taken for (\ref{eq:phiI}),  then,
\begin{align}
d_{\text{EMD}} (S_1, S_{\textbf{OA}}^{(\lambda_1)}) = \frac{1}{N} \sum_{i} \| x_i - u_{\phi_I(i)} \|_2,
\end{align}
which in turn means that $\phi_I$ is the optimal assignment from  $S_1$ to $S_{\textbf{OA}}^{(\lambda_1)}$ by the definition of the EMD. So the first intermediate argument is proved.

The second intermediate argument is that $\phi^*$ is the optimal assignment from   $S_{\textbf{OA}}^{(\lambda_1)}$ to $S_2$. This argument can be proved samely as the first one. Say the equal sign being taken for (\ref{eq:ineq2}) is equivalent to that
\begin{equation}
    d_{\text{EMD}} ( S_{\textbf{OA}}^{(\lambda_1)}, S_2) = \frac{1}{N} \sum_{i} \| u_i- y_{\phi^*(i)} \|_2.\vspace{-0.3em}
\end{equation}
Thus, $\phi^*$ is the optimal assignment from   $S_{\textbf{OA}}^{(\lambda_1)}$ to $S_2$ is proved.

Then, with the two intermediate arguments, we can reformalize the setup to regard that  $S_{\textbf{OA}}^{(\lambda_2)}$  is interpolated from source pairs $S_{\textbf{OA}}^{(\lambda_1)}$ and $S_2$ with the mix ratio $\frac{\lambda_2 - \lambda_1}{1-\lambda_1}$, because the optimal assignment from $S_{\textbf{OA}}^{(\lambda_1)}$ to $S_2$ is the same as the optimal assignment from $S_1$ to $S_2$. This argument then becomes an isomorphic with respect to the first intermediate argument. Then we prove that $\phi_I$ is the optimal assignment from $S_{\textbf{OA}}^{(\lambda_1)}$ to $S_{\textbf{OA}}^{(\lambda_2)}$ similarly as the proof for the first intermediate argument.
\end{proof2}

\begin{proof3}
We have shown that $\phi_I$ is optimal assignment between $S_{\textbf{OA}}^{(\lambda_1)} = \{u_k\} =\{  (1-\lambda_1) \cdot x_k + \lambda_1 \cdot y_{\phi^*(k)}\} $ and $S_{\textbf{OA}}^{(\lambda_2)} = \{v_l\} =\{  (1-\lambda_2) \cdot x_l + \lambda_2 \cdot y_{\phi^*(l)}\} $. Thus,
$d_\text{EMD}(S_{\textbf{OA}}^{(\lambda_1)}, S_{\textbf{OA}}^{(\lambda_2)})    =  \frac{1}{N} \sum_{k} \|  ( (1-\lambda_1) \cdot x_k + \lambda_1 \cdot y_{\phi^*(k)} ) - ((1-\lambda_2) \cdot x_{\phi_I(k)} + \lambda_2 \cdot y_{\phi^*(\phi_I(k))}) \|_2 =  \frac{1}{N} \sum_{k} \|(\lambda_2 - \lambda_1) (x_k - y_{\phi^*(k)} ) \|_2 = (\lambda_2 - \lambda_1)  \frac{1}{N} \sum_{k} \|(x_k - y_{\phi^*(k)} ) \|_2 = (\lambda_2 - \lambda_1) \cdot d_\text{EMD}(S_1, S_2)$.
\end{proof3}

\begin{algorithm*}[b!]
  \caption{\textbf{Episodic training of ProtoNet with PointMixUp.} {\small From line 3 to line 8 is where PointMixUp takes a role in addition to the ProtoNet baseline. Testing stage is similar as training stage, but without line 13 and line 14 which learn new weight from query examples.} }
  \label{alg:few-shot}
  \begin{algorithmic}[1]
  \small
  \Require Set of sampled episodes $ \{\mathcal{D}_i\}$, where $\mathcal{D}_i = \mathcal{D}_i^S \cup \mathcal{D}_i^Q$ denoting the support and query sets
  \Require $h_\theta$: feature extractor network: input $\to$ latent embedding
  
  \State randomly initialize $\theta$
  
  \For{episode $i$}
    \For{class $c$}
        \State calculate prototype $\Bar{z}_c$ from $ \mathcal{D}^S_i$, with $h_\theta$.
    \EndFor
    \State {Construct Mixup samples $\mathcal{D}^\text{mix}_i$ from support set $\mathcal{D}^{S}_i$}.
    \State Predict the label distributions for mixed examples in $\mathcal{D}^\text{mix}_i$, with distance to $\Bar{z}_c$.
    \State Update $\theta$ with prediction from mixed examples, as episode-specific weights $\theta_i$.
    \For{class $c$}
        \State calculate new prototype $\Bar{z}_c^{(\theta_i)}$ from $ \mathcal{D}^S_i$, with  $h_{\theta_i}$
    \EndFor
    \State Predict the label distributions for query examples in $\mathcal{D}^Q_i$, with distance to $\Bar{z}_c^{(\theta_i)}$.
    \State Update $\theta_i$ with prediction from query examples.
    \State $\theta \gets \theta_i$
  \EndFor
  \State \Return $\theta$
  \end{algorithmic}
\end{algorithm*}

\section{Few-shot learning with PointMixUp}

We test if our PointMixup helps point cloud few-shot classification task, where
a classifier must generalize to new classes not seen in the training set, given only
a small number of examples of each new class. We take ProtoNet~\cite{snell2017protonet} as the baseline method for few-shot learning, and PointNet++~\cite{qi2017pointnet++} is the feature extractor $h_\theta$.

\subsubsection{Episodic learning setup}
ProtoNet takes the episodic training for few-shot learning, where an episode is designed to mimic the few-shot task by subsampling classes as well as data.
A $N_C$-way $N_S$-shot setting is defined as that in each episode, data from $N_C$ classes are sampled and $N_S$ examples for each class is labelled. In the $i^\text{th}$ episode of training, the dataset $\mathcal{D}_i$ consists of the training example and class pairs from $N_C$ classes sampled from all training classes. Denote $\mathcal{D}_i^S \subset \mathcal{D}_i$ is the support set which consists of labelled data from $N_C$ classes with $N_S$ examples, and $\mathcal{D}_i^Q  = \mathcal{D}_i \backslash \mathcal{D}_i^S$ is the query set which consists of unlabelled examples to be predicted. 

\subsubsection{Baseline method for few-shot classification: ProtoNet~\cite{snell2017protonet}}
In each episode  $\mathcal{D}_i$, ProtoNet computes a prototype as the mean of embedded support examples $\Bar{z}_c$ for each class $c$, from all examples from the support set $\mathcal{D}_i^S$. The latent embedding is from the network $h_\theta$ (for which we use PointNet++~\cite{qi2017pointnet++} without the last fully-connected layer). Then each example $S$ from the query set $\mathcal{D}_i^Q$ is classified into a label distribution by a softmax over (negative) distance to the class prototypes: $$p(\hat{y} = c|S) = \frac{\text{exp}(-d (S,\Bar{z}_c))}{\sum_{c'}\text{exp}(-d (S,\Bar{z}_{c'}))},$$
where $d(\cdot, \cdot)$ is the Eudlidean distance in the embedding space. 
In training stage, the weights $\theta$ for the feature extractor $h_\theta$ is updated by the cross-entropy loss for the predicted query label distribution and the ground truth.

\subsubsection{Few-shot point cloud classification with PointMixup}
We use PointMixup to learn a better embedding space for each episode.
Instead of using the $h_\theta$ directly to predict examples from query set, we learn a episode-specific weight $\theta_i$ from the mixed data, and the query examples are predicted by $h_{\theta_i}$.
We use PointMixup to construct a mixed set  $\mathcal{D}_i^\text{mix}$ from the labelled support set $\mathcal{D}_i^S$, which consists of examples from $\binom{N_c}{2}$ class pairs and for each class pairs $N_s$ mixed examples are constructed from randomly sampling support examples. Then the weight $\theta$ is updated as $\theta_i$ from backprop the loss from the prediction of mixed examples from  $\mathcal{D}_i^\text{mix}$. After that, the label of query examples from  $\mathcal{D}_i^Q$ is then predicted with the updated feature extractor $h_{\theta_i}$. See Algorithm~\ref{alg:few-shot} for an illustration of the learning scheme.

\section{Further Discussion on Interpolation Variants}
The proposed PointMixUp adopts \emph{Optimal Assignment (OA) interpolation} for point cloud because of its advantages in theory and in practice. To compare Optimal Assignment interpolation with the two alternative strategies, \emph{ Random Assignment (RA) interpolation} and \emph{Point Sampling (PS) interpolation}, 
the proposed PointMixUp with OA interpolation is the best performing strategy, followed by PS interpolation. RA interpolation, which has a non-shortest path definition of interpolation, does not perform well.

Here we extend the discussion on the two alternative interpolation strategies, through which we analyze the possible advantages and limitations under certain conditions, which in turn validates our choice of applying Optimal Assignment interpolation  for PointMixup.

\subsubsection{Random Assignment interpolation}
From our shortest path interpolation hypothesis for Mixup, the inferiority of RA interpolation comes from that it does not obey the shortest path interpolation rule, so that the mixed point clouds from different source examples can easily entangle with each other. From Fig. 3 in the main paper, the Random assignment interpolation produces chaotic mixed examples which can hardly been recognized with the feature from the source class point clouds. Thus, RA interpolation fails especially under heavy Mixup (the value of $\lambda$ is large).

\subsubsection{Point Sampling interpolation: yet another shortest path interpolation}
Point Sampling interpolation performs relatively well in PointNet++ and sometimes comparable with the Optimal Assignment interpolation. From Fig. 3 in the main paper, the PS interpolation produces mixed examples which can be recognized which classes of source data it comes from.

Reviewing the shortest path interpolation hypothesis, We argue that when the number of points $N$ is large enough, or say $N \to \infty$, Point Sampling interpolation also (approximately) defines a shortest path on the metric space $(\mathcal{S}, d_\text{EMD})$ (Note that given the initial and the final points, the shortest path in $(\mathcal{S}, d_\text{EMD})$ is not unique). This is a bit counter-intuitive, but reasonable.

We show the \textit{shortest path property}. 
Recall that point sampling interpolation randomly draws without replacement of points from each set are made according to the sampling frequency $\lambda$: $S_{\textbf{PS}}^{(\lambda)} = S_1^{(1-\lambda)} \cup S_2^{(\lambda)},$
where $S_2^{(\lambda)}$ denotes a randomly sampled subset of $S_2$, with $ \lfloor \lambda N \rfloor$ elements. ($\lfloor \cdot \rfloor$ is the floor function.) And similar for $S_1^{(1-\lambda)}$ with $N - \lfloor \lambda N \rfloor$ elements, such that $S_{\textbf{PS}}^{(\lambda)}$ contains exactly $N$ points.
Imagine that a subset $S_1^{(1-\lambda)}$ with a number of  $N - \lfloor \lambda N \rfloor$  points in $S^{(\lambda)}_\textbf{PS}$ are identical with that in $S_1$. For $d_\text{EMD} (S^{(\lambda)}_\textbf{PS}, S_1)$, the optimal assignment will return these identical points as matched pairs, thus they contribute zero to the overall EMD distance. Thus, 
\begin{align*}
d_\text{EMD} (S^{(\lambda)}_\textbf{PS}, S_1) &= \frac{N - \lfloor\lambda N \rfloor}{N} d_\text{EMD} (S^{(\lambda)}_\textbf{PS} \setminus S_1^{(1-\lambda)}, S_1\setminus S_1^{(1-\lambda)})\\
&= \frac{N-\lfloor\lambda N \rfloor}{N} d_\text{EMD} (S_2^{(\lambda)}, S_1 \setminus S_1^{(1-\lambda)})\\
& \approx \frac{N-\lfloor\lambda N \rfloor}{N}  d_\text{EMD} (S_2 , S_1)\\
&\approx (1-\lambda) \cdot  d_\text{EMD} (S_1 , S_2),
\end{align*}
from which $d_\text{EMD} (S_2^{(\lambda)}, S_1 \setminus S_1^{(1-\lambda)}) \approx  d_\text{EMD} (S_2 , S_1)$ is because that $S_1$ and $S_1 \setminus S_1^{(1-\lambda)}$ are the point clouds representing the same shape but with different density, and the same with $S_2$ and $S_2^{(\lambda)}$.

Similarly, $d_\text{EMD} (S^{(\lambda)}_\textbf{PS}, S_2) \approx \lambda \cdot  d_\text{EMD} (S_1 , S_2)$, and thus $d_\text{EMD} (S^{(\lambda)}_\textbf{PS}, S_1) + d_\text{EMD} (S^{(\lambda)}_\textbf{PS}, S_2) = d_\text{EMD} (S_1, S_2)$, which in turn proves the shortest path property.

We note that the \textit{linearity} of PS interpolation w.r.t. $d_\text{EMD}$ also holds and the proof can be derived similarly. Thus, although strictly not an ideally continuous interpolation path, PS interpolation is (appoximately) a shortest path linear interpolation in $(\mathcal{S},d_\text{EMD})$, which explains its good performance.

\subsubsection{Point Sampling interpolation: limitations}

The limitation of PS interpolation is from that the mix ratio $\lambda$ controls change of local density distribution, but the underlying shape does not vary with $\lambda$. So, as shown in Table~\ref{tab:pointnet}, PS interpolation fails with PointNet~\cite{qi2017pointnet}, which is ideally invariant to the point density, because a max pooling operation aggregates the information from all the points.

\begin{table}[t!]
\centering
\caption{\textbf{Different interpolation strategies on PointNet~\cite{qi2017pointnet}} Following the original paper~\cite{qi2017pointnet} we test on unaligned setting. PS interpolation fails with PointNet as a density-invariant model. The numbers are accuracy in percentage.}
\begin{tabular}{@{}c|cccc@{}}
\toprule
          \hspace{0.5em}{Baseline} \hspace{0.5em} &  \hspace{0.5em}{PointMixup} \hspace{0.5em} &   \hspace{0.5em}{Random Assignment} \hspace{0.5em} & \hspace{0.5em} {Point Sampling}  \hspace{0.5em}\\ \hline
   89.2       &{\bf  89.9}   & 88.2    & 88.7     \\\bottomrule
\end{tabular}
\label{tab:pointnet}
\end{table}

A question which may come with PS interpolation is that how it performs relatively well with PointNet++, which is also designed to be density-invariant. This is due to the sampling and grouping stage. PointNet++ takes same operation as PointNet in learning features, but in order to be hierarchical, the sampling and grouping stage, especially the farthest point sampling (fps) operation is not invariant to local density changes such that it samples different groups of farthest points, resulting in different latent point cloud feature representations. Thus, PointNet++ is invariant to global density but not invariant to local density differences, which makes PS interpolation as a working strategy for PointNet++. However, we may still expect that the performance of Mixup based on PS interpolation is limited, because it does not work well with PointNet as a basic component in PointNet++. 

By contrast, the proposed PointMixup with OA interpolation strategy is not limited by the point density invariance. As a well established interpolation, OA interpolation smoothly morphes the underlying shape. So we claim that OA interpolation is a more generalizable strategy.

\input{6-supp.bbl}